\documentclass[fleqn,10pt,twocolumn]{AROB}

\title{Proposition of Affordance-Driven Environment Recognition Framework\\
Using Symbol Networks in Large Language Models}

\author{Kazuma Arii${}^{1\dagger}$ and Satoshi Kurihara${}^{2}$}
\speaker{Kazuma Arii}

\affils{${}^{1}$Department of Science and Technology, Keio University, Kanagawa, Japan\\
(E-mail: arkzm1024@keio.jp)\\
${}^{2}$Faculty of Science and Technology, Keio University, Kanagawa, Japan\\
(E-mail: satoshi@keio.jp)\\
}
\abstract{
In the quest to enable robots to coexist with humans, understanding dynamic situations and selecting appropriate actions based on common sense and affordances are essential. Conventional AI systems face challenges in applying affordance, as it represents implicit knowledge derived from common sense. However, large language models (LLMs) offer new opportunities due to their ability to process extensive human knowledge. This study proposes a method for automatic affordance acquisition by leveraging LLM outputs. The process involves generating text using LLMs, reconstructing the output into a symbol network using morphological and dependency analysis, and calculating affordances based on network distances. Experiments using ``apple'' as an example demonstrated the method's ability to extract context-dependent affordances with high explainability. The results suggest that the proposed symbol network, reconstructed from LLM outputs, enables robots to interpret affordances effectively, thereby bridging the gap between symbolized data and human-like situational understanding.}

\keywords{%
Knowledge Graph, Affordance, Large Language Models
}

\begin{document}
\maketitle

\section{Introduction}
The concept of affordances, originally introduced by Gibson (1979)\cite{gibson1979}, emphasizes the actionable possibilities offered by objects or environments in relation to an agent's capabilities. This foundational theory has since influenced various domains, including robotics, computer vision, and interaction design. In the study of autonomous agents such as robots, the acquisition of affordances depending on the situation plays an important role. For example, when an agent perceives an apple and a knife, it can recall actions such as ``use the knife to cut the apple'', and when it observes an apple in a store, it can recall actions such as ``buy an apple''. If it were possible to acquire affordances like ``buy an apple'' when an agent observes an apple at a shop, the search space for planning could be significantly reduced. 

However, since affordance is implicit knowledge derived from common sense, it is necessary to gather the knowledge and extract the affordance from it for mechanical affordance acquisition. Commonsense reasoning is required for many intellectual tasks, including natural language processing, computer vision, planning  and so on. 

However, since affordance is implicit knowledge derived from common sense, it is necessary to gather knowledge and extract affordance for mechanical affordance acquisition. Commonsense reasoning is essential for many intellectual tasks, including natural language processing, computer vision, and planning\cite{davis2015commonsense}, and so on. In particular, robotics offers a practical domain where affordance learning can be directly applied, enabling robots to understand object properties and their uses, and to learn behaviors in physical environments. For example, Fitzpatrick et al. (2003)\cite{Fitzpatrick2003} explored how robots can learn affordances by interacting with objects. Similarly, Montesano et al. (2008)\cite{Montesano2008} proposed a model for learning object affordances through the coordination of sensory and motor activities, extending this understanding to imitation-based tasks. These studies highlight the critical role of affordance learning in advancing robotic intelligence. 

In light of this background, we have developed three affordance acquisition methods: one to extract human knowledge from a large language model (LLM), one to structure it as a knowledge graph, and one to acquire affordance from the knowledge graph. This approach enables the acquisition of affordances that are close to those of human beings and that accurately reflect the current situation.

In this paper, we describe the knowledge base for mechanically handling common sense, including tacit knowledge, and the ontology learning involving LLM in Section 2. In Section 3, we present the three methods for acquiring affordances. We report the results of experiments on the acquisition of affordances using the constructed network and compare them with those by humans in Section 4. We conclude in Section 5 with a brief summary and a discussion of the remaining issues.

\section{Related Works}
The acquisition of affordances, which refers to understanding the actionable possibilities an environment offers to an agent, often involves reasoning about the relationships between entities, objects, and actions. Knowledge bases and ontology learning methods provide structured frameworks for encoding these relationships, offering a foundation for affordance discovery and representation.

\subsection{Knowledge Base}
Up to now, most research in the field has focused on how to construct a knowledge base, which is a database of human common knowledge. In this section, we discuss some of the efforts that have been made to date in the area of knowledge bases.

WordNet \cite{miller1995wordnet} is an online lexical database of the English language created to make the semantic relations of lexemes available in a machine-readable way. WordNet consists of synonym sets of nouns, verbs, adjectives, and adverbs. Each set defines a lexical concept corresponding to its meaning. It contains more than 166,000 word forms, and semantic relations between pairs of lexical concepts.

DBpedia \cite{auer2007dbpedia} is a community project that provides a rich corpus of diverse data and gives researchers the opportunity to develop, compare, and evaluate extraction, inference, and uncertainty management methods and deploy operational systems on the Semantic Web.

The main contributions of DBpedia are the development of an information extraction framework that converts Wikipedia content into RDF format, providing foundational components for advanced research in areas such as information extraction, clustering, uncertainty management, and query processing. DBpedia also offers Wikipedia's content as an extensive multi-domain RDF dataset, enabling its use in various Semantic Web applications. Furthermore, the interconnection of the DBpedia dataset with other public datasets has resulted in a massive data web containing approximately 2 billion RDF triples. To enhance usability, several interfaces and modules have been developed to provide Web-based access to the dataset and support integration with other sites.

Knowledge graphs have been extensively studied as tools to model and reason about relationships between objects. Wang et al. (2014)\cite{Wang2014} proposed a method for embedding knowledge graphs by translating on hyperplanes, enabling efficient reasoning over multi-relational data structures and facilitating the inference of relationships between entities.

\subsection{Affordances Detection}
Nguyen et al. (2016)\cite{Nguyen2016} introduced a convolutional neural network-based approach for detecting object affordances directly from images, integrating object detection and affordance recognition within a unified framework. Building on this, Do et al. (2018) proposed AffordanceNet\cite{Do2018} , an end-to-end deep learning model that jointly performs object segmentation and affordance map generation, achieving significant improvements in accuracy and efficiency.

\subsection{Affordance with LLMs}
Recent advancements in artificial intelligence have highlighted the potential of LLMs to revolutionize ontology learning, a critical area in knowledge representation and extraction. Giglou et al. (2023) proposed the LLMs4OL paradigm \cite{babaei2023llms4ol}, which systematically explores the application of LLMs for ontology learning tasks such as term typing, taxonomy discovery, and non-taxonomic relation extraction. Their comprehensive evaluation of state-of-the-art LLMs (e.g., GPT-4, Flan-T5) on diverse datasets (WordNet, GeoNames, UMLS) underscores the promise and challenges of leveraging these models in automated ontology construction. Their study further emphasizes the role of fine-tuning in improving LLM performance and sets the stage for future research integrating domain expertise and AI methods in ontology development.

Adak et al. (2024)\cite{Adak2024} introduce the TEXT2AFFORD dataset, a novel crowdsourced resource designed to evaluate the ability of language models to predict contextual object affordances. Their findings reveal significant limitations in current models, including state-of-the-art LLMs and VLMs, in reasoning about affordances solely from text. This study highlights the importance of grounding for improving physical reasoning capabilities and emphasizes the challenges of understanding affordances in the absence of visual context.

\section{Methods}
Recent advances in integrating LLMs with robotic systems have shown promise in enabling robots to execute high-level natural language instructions. A notable contribution in this area is the SayCan framework, introduced by Ahn et al. (2022) \cite{ahn2022icanisay}. SayCan combines the semantic reasoning capabilities of LLMs with robots' pretrained affordance-based skills to bridge the gap between abstract language instructions and actionable robotic tasks. By leveraging LLMs for high-level task planning and pretrained value functions for grounding feasibility in real-world environments, SayCan allows robots to perform complex, temporally extended tasks, such as multi-step manipulation and navigation in dynamic settings like kitchens.

This approach has demonstrated the successful synergy of symbolic reasoning and embodied capabilities, achieving state-of-the-art results on 101 real-world tasks. Notably, SayCan highlights the scalability of robotic systems with the continued improvement of LLMs, as upgrading the language model significantly enhances task execution performance. The framework also provides interpretability by generating a stepwise breakdown of tasks, a feature critical for understanding and refining robot behavior in human-centric environments.

Our proposed approach comprises three methods: one for dataset generation, one for knowledge graph construction, and one for affordance acquisition. These methods are utilized to generate sentences from LLMs to create a dataset, construct a knowledge graph from the generated sentences using a natural language processor, and obtain the affordance acquisition from the constructed knowledge graph, respectively.

\subsection{Dataset Generation}
LLMs, which learn a large amount of text in advance and incorporate a vast amount of human knowledge in the learning process, can acquire sentences that contain affordances. However, in terms of the model's process and basis of inference, an LLM is a black box, and the form of the output is not constant. Therefore, by analyzing the output sentences of an LLM, it is possible to symbolize and extract human knowledge. This makes it possible to restrict the output to specific types of symbols to increase generality and visualize the behavior when applying the obtained affordances to downstream tasks.

In this study, we use GPT-4 Turbo (GPT) as the LLM due to its simplicity and low cost.

Also, we aim at acquiring affordances under specific conditions among the knowledge possessed by humans. In this case, the subject of the output sentence is guided by a prompt to be ``I'' because, to acquire knowledge without commonsense, the subject of the output sentence should be an action that acts on an object. Another reason is that we want to avoid generating sentences in which a non-human subject is a subject, since this would result in the output of actions that a human would not choose to do or would be unlikely to perform.

Furthermore, it is possible to acquire a wider range of knowledge by collecting knowledge about objects that include location information, such as ``apple on table'', and knowledge about attributes associated with objects and actions, such as ``red'' and ``in kitchen''.

We collect knowledge about a specific one-word object and construct a network. First, the prompt is given 200 times to output a sentence with the object of the object to be collected. Next, the GPT is given ten prompts for outputting objects that contain location information and then ten prompts for outputting actions with the corresponding noun phrase as the object for each element of the set of objects in the collected knowledge, excluding the original object. Finally, the GPT is prompted an additional ten times to gather knowledge about the collected attributes.

\subsection{Knowledge Graph Construction}
In this section, we explain our method for generating a knowledge network by performing morphological and dependency parsing on sentences acquired from LLM. We use Stanford CoreNLP \cite{manning2014corenlp} as the morphological and dependency parser.

\subsubsection{Node Composition}
\label{sec:node_composition}
\begin{figure}[b]
 \begin{center}
 \includegraphics[width=8cm]{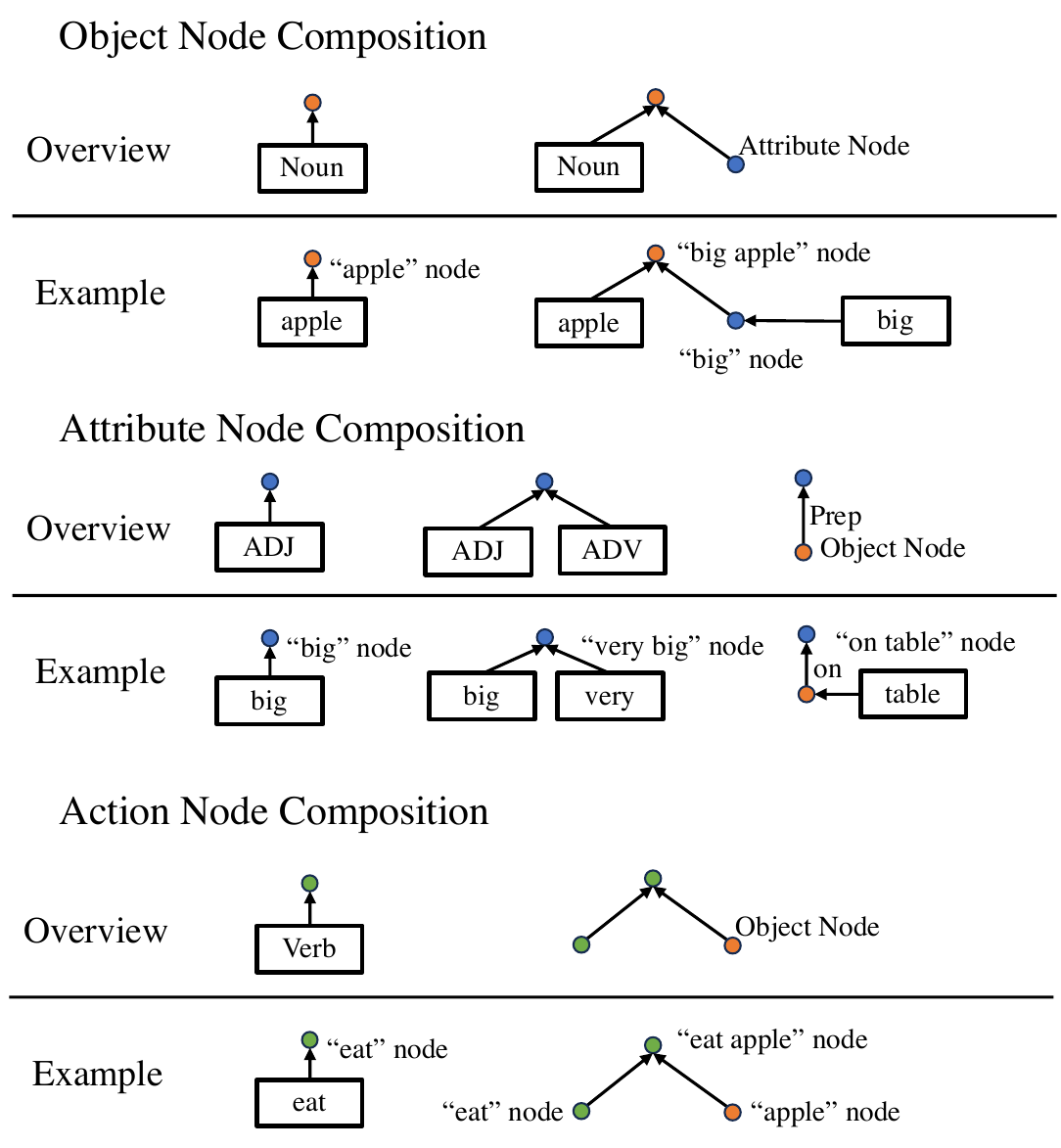}
\caption{\label{fig:node_composition} How to compose the nodes of the knowledge graph from sentences. The nodes surrounded in boxes are Origin Nodes (formal nodes: nodes for composition of other nodes.) }
 \end{center}
\end{figure}

Nodes consist of an Object Node for objects, an Attribute Node for attributes, an Action Node for actions, and an Origin Node for the composition of each node. An overview of the composition method for each node except the Origin Node is shown in Fig. \ref{fig:node_composition}.

The nodes surrounded by boxes in the figure represent Origin Nodes, each of which is composed of other nodes; Object Nodes, which correspond to an unmodified object and an object with modifying attributes (from left to right in the figure); Attribute Nodes, which correspond to an adjective, an adjective with adverb modification, and a noun-modifying phrase with a preposition, (from left to right); and Action Nodes, which correspond to an action without an object and an action with an object (from left to right).

\subsubsection{Edge Composition}
\label{sec:edge_composition}
\begin{figure}[bt]
 \begin{center}
 \includegraphics[width=8cm]{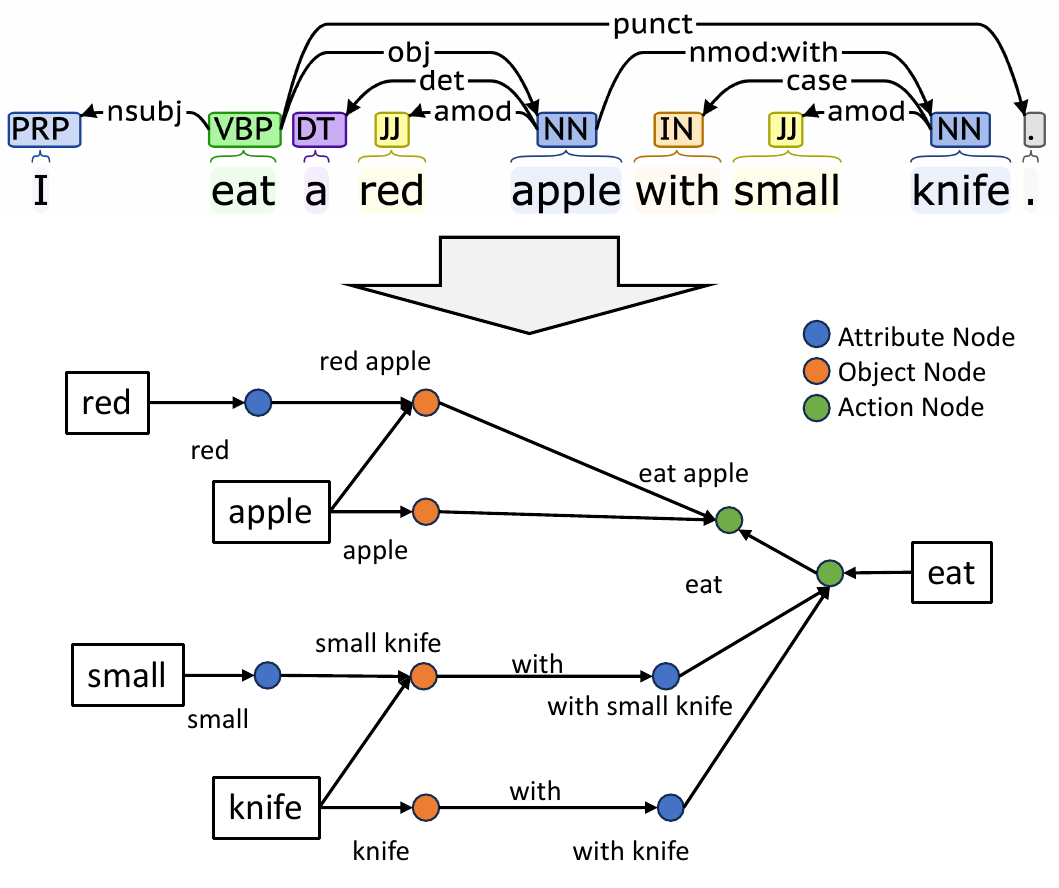}
\caption{\label{fig:edgeConstruction} An example of edge composition for network construction from a sentence (whose dependencies are analyzed by CoreNLP). If a noun has any modifier, the original noun and the modified noun are constructed as separate nodes. The nodes that make up the object are connected to its Action Node, and the nodes that make up the modifier district are connected to its Attribute Nodes.}
 \end{center}
\end{figure}

In order to obtain affordances from the environment, we connect the nodes constructed in this section to each other, as shown in Fig. \ref{fig:edgeConstruction}. This corresponds to recalling actions from objects and environments.

As an action recall from an object, an edge is connected from an Object Node to an Action Node corresponding to an action that takes the thing as its object. In this case, we use an Action Node whose Object Node is the object of the action. This makes it possible to acquire affordances for objects by searching for verbs that originate from objects. Also, the edge is connected from the Attribute Node to the Action Node as the recall of the action from the environment of the action. At this time, the Action Node is assumed to be purposeless. This makes it possible to acquire affordances by performing a search that starts from the environment when considering affordances from the environment and tools.

\subsection{Affordance Acquisition}
The constructed graph is weighted based on the number of edges in its construction. The weight of the edge connecting node $s$ to node $e$, $distance(s,e)$, is defined by 

\begin{eqnarray}
 \begin{array}{rcl}
 distance(s,e) = decay^{n}
 \end{array}
 \label{eq:distance}
\end{eqnarray}

where $decay$ is a constant satisfying $0<decay<1$ and $n$ is the number of its composition.

The power of the action of the affordance ($a$ in the graph) from the object/attribute ($x$ in the graph) is defined as $affordance(x,a)$. It is calculated as the shortest distance from $x$ to $a$ if such a path exists, or it gives a sufficiently large constant $penalty$.

The more times the combination of objects and actions appear, the closer they are expected to be to each other on the network. Therefore, the less the value of affordance is, the more the power of the action's affordance is.

\section{Results}
In this section, we construct a network utilizing the method described above. The noun ``apple'' is given as the first word and the network is then constructed. When obtaining affordances, we set $decay=0.99,\ penalty=5.0$.

In this research, we generate a total of 19,650 sentences and collect 390,816 sentences. A network is then constructed on the basis of the generated sentences. In the following, we describe the experiment we conducted on the affordances obtained on this network.

\subsection{Affordance Acquisition}
We experimentally investigate how affordances are output on the network. The network we constructed mainly collects the word ``apple'', and we check which affordances are obtained from this environment . First, we obtain affordances when an apple is observed. We assume that the apple corresponds to the Object Node ``apple'' on the network, and acquired affordances from that node. The top 10 affordances from the acquired affordances are listed in Table \ref{tab:apple_affordance}, and the top 5 affordances obtained from various environments including ``apple'' are listed in Table \ref{tab:variable_affordance}.

\begin{table}[tb]
 \centering
 \begin{tabular}{ll|ll}\hline
 action & affordance & action& affordance \\
 \hline
 choose apple & 0.60e-8 &
 leave apple & 0.000015\\
 place apple & 0.90e-9 &
 slice apple & 0.000015\\
 use apple & 0.16e-6 &
 share apple & 0.000022\\
 find apple & 0.27e-6 &
 see apple & 0.000047\\
 pick apple & 0.23e-5& 
 toss apple & 0.000055\\
 \hline
 \end{tabular}
\caption{Affordances from ``apple'' Object Node showing the top 10 most recalled actions. The actions are represented as Action Nodes in the network. The smaller the affordance value, the shorter the distance on the network, and the stronger the corresponding recalled action.}\label{tab:apple_affordance}
\end{table}

\begin{table}[tb]
	\centering
	\resizebox{\linewidth}{!}{
		\begin{tabular}{ll|ll}\hline

 \multicolumn{2}{c|}{``fall apple''} & \multicolumn{2}{c}{``sweet apple''}\\
 action & affordance &action & affordance\\
 \hline
 pick up apple & 0.94 & taste apple & 0.79 \\
 toss apple & 0.94 & smell apple & 0.85 \\
 kick apple & 0.96 & choose apple & 0.86 \\
 place apple & 0.96 & crave apple & 0.87 \\
 gather apple & 0.97 & enjoy apple & 0.87 \\
 \hline
 \multicolumn{2}{c|}{``apple'' + ``at store''} & \multicolumn{2}{c}{``apple'' + ``in tree''}\\
 action & \multicolumn{1}{l|}{affordance} & \multicolumn{1}{l}{action} & \multicolumn{1}{l}{affordance}\\
 \hline
 buy apple & 0.076 & pick apple & 0.47 \\
 find apple & 0.32 & place apple & 0.48 \\
 see apple & 0.37 & find apple & 0.48\\
 compare apple & 0.39 & compare apple & 0.48\\
 choose apple & 0.43 & photograph apple & 0.49\\
 \hline
 
 \multicolumn{2}{c|}{``apple'' + ``with friend''} & \multicolumn{2}{c}{``apple'' + ``knife''}\\
 \multicolumn{1}{l}{action} & \multicolumn{1}{l|}{affordance} & \multicolumn{1}{l}{action} & \multicolumn{1}{l}{affordance}\\
 \hline
 share apple & 0.000026 & slice apple & 0.33\\
			trade & 0.36 & cut apple & 0.35\\
			trade apple & 0.36 & peel apple & 0.39\\
			enjoy apple & 0.41 & pack apple & 0.48\\
			share  & 0.42 & share apple & 0.50\\
 \hline
			\multicolumn{2}{c|}{``apple'' + ``pencil''} & \multicolumn{2}{c}{``apple'' + ``camera''}\\
			\multicolumn{1}{c}{action} & \multicolumn{1}{c|}{affordance} &\multicolumn{1}{c}{action} & \multicolumn{1}{c}{affordance}\\
			\hline
			draw picture & 0.98 & photograph apple & 0.75\\
			sketch apple & 1.00 & use apple & 0.82\\
			trade pencil & 1.07 & take apple & 0.83\\
			find pencil & 1.23 & have apple & 0.83\\
			draw & 1.26 & take photo & 0.91\\
			\hline
		\end{tabular}
	}
 
	\caption{Affordances from different environments. The smaller the affordance value, the shorter the distance on the network, and the stronger the corresponding recalled action.}
	\label{tab:variable_affordance}
\end{table}

In the following, we compare the affordances in different environments. The strength of affordance is determined by the co-occurrences, which is a function of the number of vocabulary occurrences. Since this makes comparing the strength of affordances in different situations difficult, we focus on the type and rank of the actions recalled by the affordances and compare them.

The objects corresponding to noun phrases that include noun modifiers have more information (attributes) than those that do not, and therefore, actions that take these attributes into account can be recalled. As shown in Table \ref{tab:variable_affordance}, ``fall apple'' evokes actions such as ``pick up apple'', ``kick apple'', and ``gather apple'', and ``sweet apple'' evokes actions such as ``taste apple'' and ``smell apple''. It can be confirmed that actions are recalled in accordance with the attributes of the objects.

Next, we examine the affordances obtained when ``apple'' is observed in different environments. As shown in Table \ref{tab:variable_affordance}, when ``apple'' is observed ``at store'', the action ``buy apple'' is recalled more strongly, and when ``with friend'', the action ``share apple'' is recalled more strongly, while when ``in tree'', the action ``share apple'' is recalled more strongly. When observed with ``in tree'', ``pick apple'' is more strongly recalled than when simply observing ``apple''. This indicates that the affordances acquired change depending on the environment, and that they are more similar to those acquired by humans.

\begin{figure}[tb]
 \begin{center}
 \includegraphics[width=8cm]{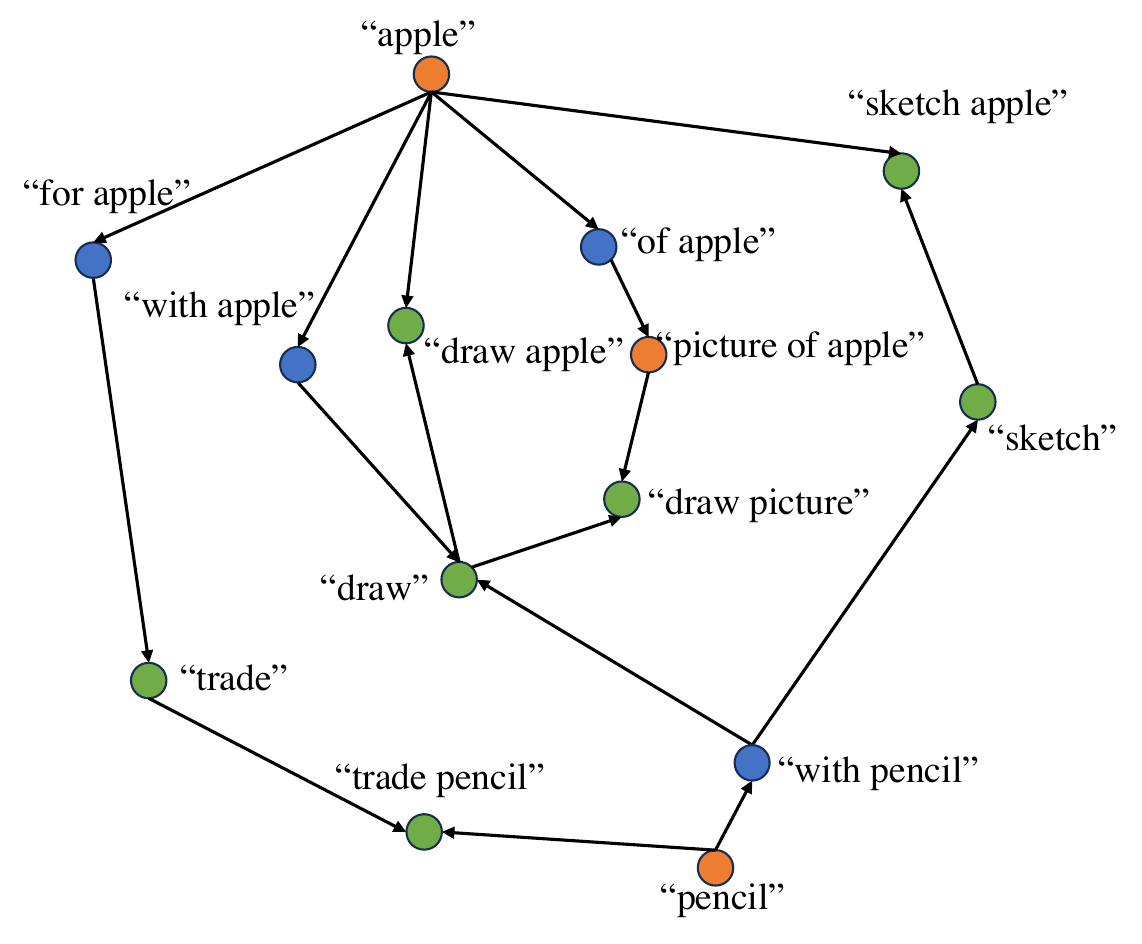}
\caption{\label{fig:exploreApplePencil} A part of the explored network when ``apple'' and ``pencil'' are observed. The distances are omitted. As shown, the actions using the pen as a tool are reached via ``with pencil'' node. It is possible to interpret this as an automatic selection of what will become a tool in the search process.}
 \end{center}
\end{figure}

Humans are able to appropriately extract and select affordances of which objects to use as tools depending on the environment. We check whether affordance acquisition that automatically selects affordance as a tool is actually performed on the constructed network. As shown in Table \ref{tab:variable_affordance}, when ``apple" and ``knife" are observed, actions such as ``slice apple" and ``cut apple" are selected, and when ``camera" is observed, actions such as ``photograph apple" and ``take photo" are recalled. In addition, when ``pencil" is observed, actions such as ``draw picture" and ``sketch apple" are recalled. These actions are recalled via ``with xxx", as shown in Fig. \ref{fig:exploreApplePencil}, for example, indicating that the tool for the action is automatically determined. These results confirm that affordances change depending on the observed objects when multiple objects are observed, and that the tools are automatically determined.

\subsection{Comparison with Humans}
Finally, we conduct a comparative experiment with humans regarding the acquisition of affordances. Specifically, we ask humans to indicate which actions they recall when given similar symbolic information and compare the coverage and its order in this study.

First, we will experiment with the coverage of human affordances.
The experimental procedure involved presenting participants with symbolic information (text) corresponding to observed objects or attributes, similar to the inputs provided to the proposed system. Participants were then asked to report the actions they could envision based on the given information. Participants were instructed to specify the action and the corresponding object when the envisioned action involved an object. The proportion of actions mentioned by participants that matched the affordances acquired by the system was calculated as a score. A weighted score was then derived by adjusting this score based on the frequency of each reported action. The results of this analysis are presented below. It should be noted that an "affordance is considered acquired" when the affordance value is less than or equal to a predefined threshold of 2.0. This threshold was determined as a value indicative of actions being recalled based on at least two observed elements, suggesting that these elements collectively contribute to the recall of the action.

\begin{table}[tb]
    \centering
    \begin{tabular}{l|ll}\hline
    observed factors & score & weighted score \\
    \hline
    ``apple'' & 75.0 & 83.3\\
    ``fall apple'' & 36.3 & 29.4 \\
    ``apple'' + ``at store'' & 66.7 & 77.8\\
    ``apple'' + ``knife'' & 50.0 & 76.9\\
    \hline
    \end{tabular}
   \caption{Coverage of human-recalled actions by affordances acquired from the network. The score is the coverage (\%) for all responses. The weighted score is calculated as the percentage of coverage for multiple respondents who have overlapping responses, treating them as separate responses.}\label{tab:compare_apple}
\end{table}

It can be seen that the weighted scores are higher except when ``fall apple'' is observed. This confirms that the weighted scores cover well the more frequently recalled actions among those recalled by humans.
The low coverage in observein``fall apple'' is also noticeable, but this can be attributed to the fact that B was generated only 37 times in total in the text, suggesting that the knowledge accumulation is insufficient. This could be attributed to the lack of frequency of sentence generation, which could be solved by generating more sentences.

Next, we compare it to the order of human affordances.We asked the participants to rank the top five movements acquired from the network and calculated the average distance from the rank. The same questions were also asked of the GPT-4o and compared to the human rankings. The results are shown in Table \ref{tab:compare_order}.

\begin{table}[b]
    \centering
    \begin{tabular}{l|ll}\hline
    observed factors & ours & GPT-4o \\
    \hline
    ``apple'' & 10.8 & 6.0\\
    ``fall apple'' & 6.6 & 6.0 \\
    ``apple'' + ``at store'' & 7.6 & 11.2\\
    ``apple'' + ``knife'' & 5.2 & 3.2\\
    \hline
    \end{tabular}
   \caption{Average of the distance of the top 5 acquired movement rankings from the human-assigned rankings. The smaller the value, the closer it is to the human-added order, indicating higher performance.}\label{tab:compare_order}
\end{table}

It can be seen that the accuracy is better when other factors are taken into account than when the “apple” is simply observed. This confirms that the more information we have, the more human-like the behavior we recall. In particular, when compared to the method of asking GPT-4o directly, it can be seen that the accuracy is comparable in many cases as the number of observation factors increases, and especially in the case of obserbing "apple" and "at store", the accuracy is remarkable. This is considered to indicate a high ability to capture information about the environment.

\section{Conclusion and Future Work}
In this paper, we proposed a series of methods for constructing a knowledge network for affordance acquisition using LLMs. The proposed approach automatically selects tools and objects for actions from the network search process and then uses them to obtain affordances according to the situation. Compared to existing methods, our proposed methods are superior in that they can acquire affordances by taking into account information associated with objects, such as word modifiers, and in that they can acquire affordances from multiple factors in a fixed format.

There are two main issues that will need to be investigated in future work. First, in this study, we constructed knowledge centered on a single object. In order to cover the full extent of human knowledge, it is necessary to collect a larger amount of knowledge over a wider range, rather than this kind of biased knowledge collection. We will then need to confirm whether this network performs its function of acquiring affordances when such generic knowledge is collected. Second, the proposed methods do not provide a meaningful affordance value for each scene. This is because the affordance value is directly related to the amount of knowledge collected. There is also a problem in that when a large number of sentences are collected, the network continues to become dense. To solve these problems, we plan to define the distance between nodes based on the probability of transition from one node to another, which will make it possible to avoid network densification and enable affordances to be numerically handled in different situations. Achieving this will allow for quantitative selection of actions through thresholding, for example.

The acquisition of affordances in more generic situations will be of great significance for the realization of general-purpose artificial intelligence. This is the motivation that will inform our future work.

\section{Acknowledgment}
This work was supported by the NEDO/Technology Development Project on Next-Generation Artificial Intelligence Evolving Together With Humans as part of the development of interactive story-type content creation to support infrastructure.

\end{document}